\documentclass{article} 
\usepackage{nips13submit_e,times}

\usepackage{times}
\usepackage{url}
\usepackage{latexsym}
\usepackage{amssymb}
\usepackage{graphicx}
\usepackage{epsfig}
\usepackage{algorithm}
\usepackage{algorithmic}
\usepackage{amsmath}
\usepackage{caption}
\usepackage{subcaption}
\usepackage{epstopdf}

\title{A Deep Learning Model for Structured Outputs with High-order Interaction}

\author{
Hongyu Guo$^{\dagger}$, Xiaodan Zhu$^{\dagger}$, Martin Renqiang Min$^{\ddagger*}$\thanks{The three authors contributed equally.}\\
National Research Council of Canada, Ottawa, ON\\
\texttt{\{hongyu.guo, xiaodan.zhu\}@nrc-cnrc.gc.ca}\\
$^{\ddagger}$ NEC Labs America, Princeton, NJ 08540\\
\texttt{renqiang@nec-labs.com} 
}

\begin{document}

\maketitle

\begin{abstract}
Many real-world applications are associated with structured data, where not only input but also output has interplay. However, typical classification and regression models often lack the ability of simultaneously exploring high-order interaction within input and that within output.  In this paper, we present a deep learning model aiming to generate a powerful nonlinear functional mapping from structured input to structured output. More specifically, we propose to integrate high-order hidden units, guided discriminative pretraining, and high-order auto-encoders for this purpose. We evaluate the model with three datasets, and obtain state-of-the-art performances among competitive methods. Our current work focuses on structured output regression, which is a less explored area, although the model can be extended to handle structured label classification. 
\end{abstract}

\section{Introduction}
Problems of predicting structured output span a wide range of fields, including natural language understanding, speech processing, bioinfomatics, image processing, and computer vision, amongst others. 
Structured learning or prediction has been approached  with many different models~\cite{Bakir:2007,DBLP:journals/jiis/GuoL13,limulti,Yujia2014,Nowozin:2011}, such as graphical models \cite{Koller:2009}, large margin-based approaches \cite{Tsochantaridis:2005}, and conditional restricted Boltzmann machines \cite{Mnih:2012}. Compared with structured label classification, structured output regression is a less explored topic in both the machine learning and data mining community. Aiming at regression tasks, methods such as continuous conditional random fields~\cite{DBLP:conf/nips/QinLZWL08} have also been successfully developed. Nevertheless, a property shared by most of these previous methods is that they often make explicit and exploit certain structures in the output spaces, which is quite limited. 

The past decade has seen the great advance of deep neural networks in modeling high-order, non-linear interactions. Our work here aims to extend such success to construct nonlinear functional mapping from high-order structured input to high-order structured output. To this end, we propose a deep High-order Neural Network with Structured Output (HNNSO). 
The upper layer of the network implicitly focuses on modeling interactions among output, with a high order anto-encoder that aims to recover correlations in the predicted multiple outputs; the lower layer network contributes to capture high-order input structures, using bilinear tensor products; and the middle layer constructs a mapping from input to output. In particular, we introduce a discriminative pretraining approach to guiding the focuses of these different layers of networks.

To the best of our knowledge, our model is the first attempt to construct deep learning schemes for structured output regression with high-order interactions.
We evaluate and analyze the proposed model on multiple datasets: one from natural language understanding and two from image processing. We show state-of-the-art predictive performances of our proposed strategy in comparison to other competitive methods.

\section{High-Order Neural Models with Structured Output}

\begin{figure}[h]
  \centering
   \epsfig{file = 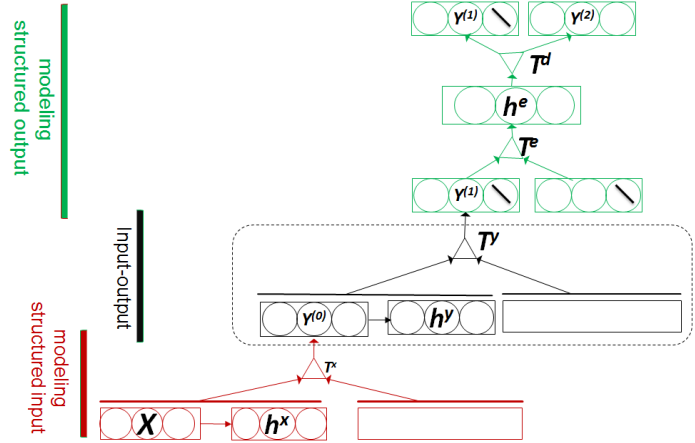, height=5.93cm}
  \caption{A specific implementation of a high-order neural network with structured output. 
  }
  \label{fig:MNTN}
\end{figure}

We regard a nonlinear mapping from structured input to structured output as consisting of three integral and complementary components in a high-order neural network. We name it as High-order Neural Network with Structured Output (HNNSO). 
Specifically, given a $D \times N$ input matrix $ [X_{1}, \dots, X_{D}]^{T}$ and a $D \times M$ output matrix $ [Y_{1}, \dots, Y_{D}]^{T}$, we aim to model the underlying mapping $f$ between the inputs $X_{d} \in \Re^{N}$ and the outputs $Y_{d} \in \Re^{M}$.  
Figure~\ref{fig:MNTN} presents a specific implementation of HNNSO. Note that other variants are allowed; for example, the dot rectangle may implement multiple  layers. 
The top layer network is a high-order de-noising auto-encoder (the green portion of Figure \ref{fig:MNTN}). In general, an auto-encoder is used for denoising input data. In our model, we use it to denoise the predicted output $y^{(1)}$ resulting from the lower layers, so as to capture the interplays among output. Similar to the strategy employed  by Memisevic in~\cite{RolandICCV2011}, 
 during training, we randomly corrupt a portion of gold labels, and the perturbed data are then fed to the auto-encoder. The hidden unit activations of the auto-encoder are first calculated by combining two versions of such corrupted gold labels, using a tensor $T^e$ to capture their multiplicative interactions.  Subsequently, the hidden layer is used to gate the top tensor $T^d$ to recover the true labels from the perturbed gold labels. As a result, 
the corrupted data force the encoder to reconstruct the true labels, in which the tensors and the hidden layer encode  the covariance patterns among the output during reconstruction. 

The bottom layer (red portion of Figure \ref{fig:MNTN}) describes a bilinear tensor-based network to multiplicatively relate  input vectors, 
  in which a third-order tensor accumulates evidence from a set of quadratic functions of the input vectors. 
In our implementation, as in~\cite{socher2013recursive}, each input vector is a concatenation of two vectors. 
Unlike~\cite{socher2013recursive}, we here concatenate two {\em dependent} vectors: the input unit $X$ ($X \in \Re^{N}$) 
 and its non-linear, first-order projected vector $h(X)$. 
Hence, the model explores the high-order multiplicative interplays not just among $X$ but also with the non-linearly projected vector $h(X)$.

We also leverage discriminative pretraining to help construct our functional mapping from structured input to structured output, in which we guide HNNSO to model the interdependency among output, among input, as well as that between input and output, where different layers of the network focus on different types of structures. 
Specifically, we pre-train the networks layer-by-layer in a bottom up fashion, using the gold output labels. The inputs to the second layer and above are the outputs of the layer right below it, except for the top layer where the corrupted gold output labels are used as input. Doing so, the bottom layer is able to focus on capturing the input structures, and the top layer can concentrate on encoding complex interaction patterns among output. Importantly, the pre-training also  makes sure that when  fine-tuning the whole networks (will be discussed later),  the inputs to the auto-encoder  have closer distributions and structured patterns as that of the true labels (as will be seen in the experimental section). 
 Consequently, the pre-train helps the auto-encoder to have  inputs with similar structures in both learning and prediction making.
   Finally, we perform fine-tuning to simultaneously optimize all the parameters of the three layers. Unlike in the  pretraining, we use the uncorrupted outputs resulting from the second layer as the input to the auto-encoder.

\textbf{Model Formulation and Learning}
As illustrated in the {\em red} portion of Figure~\ref{fig:MNTN}, HNNSO first calculates quadratic interactions among the input and its nonlinear transformation. In detail, it first computes the hidden vector from the provided input $X$. 
For simplicity, we apply a standard linear neural network layer (with weight $W^{x}$ and bias term $b^{x}$) followed by the $tanh$ transformation: 
$h^{x}=tanh(W^{x}X+b^{x}),
$
where $tanh(z) = \frac{e^{z}-e^{-z}}{e^{z}+e^{-z}}$. Next, the first layer outputs are calculated as: 
\begin{align}
Y^{(0)} = tanh
(
\left[ 
\begin{matrix}
X\\
h^{x} \\
\end{matrix} 
\right]^{T}
\mathcal{T}^{x}
\left[ 
\begin{matrix}
X\\
h^{x} \\
\end{matrix} 
\right]
+
W^{(0)}\left[ 
\begin{matrix}
X\\
h^{x} \\
\end{matrix} 
\right]
+b^{(0)}
)
\label{bilinear1100}
\end{align}
The term $
(W^{(0)}\left[ 
\begin{matrix}
X\\
h^{x} \\
\end{matrix} 
\right]
+b^{(0)})$  
 here is similar to the standard linear neural network layer. The addition term is a bilinear tensor product with a third-order tensor $\mathcal{T}^{x}$. The tensor relates two vectors, each concatenating the input unit $X$ with the learned hidden vector $h^{x}$. 
The computation for the second hidden layer $Y^{(1)}$ is similar to that of the first hidden layer $Y^{(0)}$. 
 When learning the   de-nosing auto-encoder layer ({\em green} portion of Figure~\ref{fig:MNTN}), the encoder  takes two copies of the input, namely $Y^{(1)}$, and feeds their pair-wise products into the hidden tensor, i.e., the encoding tensor $\mathcal{T}^{e}$:
\begin{align}
h^{e} = tanh
(
[Y^{(1)}]^{T}
\mathcal{T}^{e}
[Y^{(1)}]
)
\label{bilinear1}
\end{align}
Next, a hidden decoding tensor $\mathcal{T}^{d}$ is used to multiplicatively combine $h^{e}$ with the input vector $Y^{(1)}$ to reconstruct the final output $Y^{(2)}$. Through minimizing the reconstruction error, the hidden tensors are forced to learn the covariance patterns within the final output $Y^{(2)}$: 
\begin{align}
Y^{(2)} = tanh
(
[Y^{(1)}]^{T}
\mathcal{T}^{d}
[h^{e}]
)
\label{bilinear2}
\end{align}
In our study, we use an auto-encoder with tied parameters for convenience. That is, the same tensor for $\mathcal{T}^{e}$ and $\mathcal{T}^{d}$. 
Also, de-noising is applied to prevent an overcomplete hidden layer from learning the trivial identity mapping between the input and output. In the de-noising process,  the two copies of inputs are corrupted independently.  
In our implementation, 
all model parameters can be learned by gradient-based optimization. 
We minimize over all input instances ($X_{i},Y_{i}$) the sum-squared loss error (note: cross-entropy will be used for classification tasks) between the output vector on the top layer and the true label vector: 
\begin{align}
l (\theta)=\sum_{i=1}^{N} E_{i}(X_{i},Y_{i};\theta) +\lambda \left\| \theta \right\|_{2}^{2}
 \end{align}
 Also, we employ standard $L_{2}$ regularization for all the parameters, weighted by  $\lambda$. 
For our non-convex objective function here, we deploy the AdaGrad~\cite{DBLP:journals/jmlr/DuchiHS11} 
  to search for the optimal model parameters. 
\section{Experiments}
\label{sec:experimental-study}
\textbf{Baselines} \newline
We compared HNNSO's predictive performance, in terms of  Root Mean Square Error (RMSE), with six regression models: (1)  the Multi-Objective Decision Trees (MODTs)~\cite{Blockeel:1998:TIC:645527.657456,Kocev1421729}; (2) a collection of Support Vector Regression (denoted as SVM-Reg)~\cite{Smola:2004:TSV:1011935.1011939} with RBF kernel, each for one target attribute; 
(3) a traditional neural network, i.e., the Multiple Layer Perceptron (MLP) with one hidden layer and multiple output nodes; (4)  the so-called multivariate multiple regression (denoted as MultivariateReg), which takes into account the correlations among the multiple targets using a matrix computation; 
(5) an approach that stacks the MultivariateReg on top of the MLP (denoted MLP-MultivariateReg); 
 and (6) the Gaussian Conditional Random fields (Gaussian\-CRF)~\cite{DBLP:conf/pkdd/Guo13,DBLP:conf/nips/QinLZWL08,Radosavljevic:2010:CCR:1860967.1861125}, in which the outputs from a MLP were used as the CRF's node features, and the square of the distance between two target variables was modeled by an edge feature. In our experiments, all the parameters of these baselines have been  carefully tuned.  

\begin{table}[h!]
\begin{center}

\begin{tabular}{|l||c|c||c|c||c|c|}\hline 
   &\multicolumn{2}{|c|}{SSTB}  &\multicolumn{2}{|c|}{MNIST} &\multicolumn{2}{|c|}{USPS}  \\\cline{2-7}
 Methods   &RMSE& relative error &RMSE& relative error&RMSE& relative error \\
&&reduction &&reduction&&reduction\\
\hline \hline%
MODTs  &0.0567 &34.2\%&0.0739 &33.1\%&0.6487 &13.8\%\\

SVM-Reg &0.0452&17.4\% &0.0602 &17.9\% &0.5977&6.4\%\\ 
MLP & 0.0721 &48.2\%& 0.0800 &38.2\%& 0.6683 &16.3\% \\ 
MultivariateReg & 0.0614 &39.2\%& 0.1097 &54.9\%& 0.6169 &9.3\% \\
MLP-MultivariateReg & 0.0705 &47.0\%& 0.0791 &37.5\% & 0.6059 &7.7\% \\  
Gaussian-CRF & 0.0706 &47.1\%& 0.0800 &38.2\%& 0.6047 &7.5\% \\  

HNNSO  &0.0373&-&0.0494&- &0.5591&-\\ 
 \hline\end{tabular}
\end{center}
\caption{Ten-fold averaged RMSE scores of models on the SSTB, MNIST, and USPS data. The differences of HNNSO from other models are statistically significant at the 95\% significance level.
 }
  \label{tab:all}
\end{table}
\begin{figure}[h]
  \centering
   {\epsfig{file = 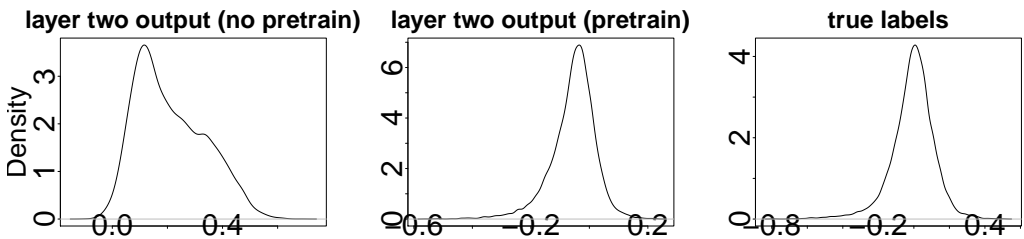, width = 14.5cm, height=3.4cm}}
  \caption{Effect of pretraining: the distributions of the predicted $Y^{(1)}s$ with pretraining (middle) were  closer to the  true labels (right), compared to the non pretrained version (left).}
  \label{fig:dist}
\end{figure}

\textbf{Datasets} \newline
There recently have been a surge of interests in using real-valued, low-dimentional vector to represent a word or a sentence in the natural language processing (NLP).
Our first experiment was set up in such a circumstance. Specifically, we used the Stanford Sentiment Tree Bank (SSTB) dataset~\cite{socher2013recursive} that contains 11,855 movie review sentences. In the best embeddings reported in ~\cite{socher2013recursive}, each sentence is represented by a 25-dimensional vector. We obtained these vectors from http://nlp.stanford.edu/sentiment/, and used the first 15 elements to predict the last 10 dimensions. 
Our second experiment used 10,000 examples from the test set of MNIST digit  database~\footnote{http://yann.lecun.com/exdb/mnist/}. 
On purpose, we employed PCA  to reduce the dimension of the data to 30, resulting in  30 PCA components that are pair-wise, linearly independent to each other. In our experiment, we used the first 15 dimensions to predict the last 15 dimensions. 
Our last experiment  used the USPS handwritten digit database~\footnote{http://www.cs.nyu.edu/~roweis/data/usps\_all.mat}.  We randomly sampled 1100 images from the original data set, and used the first half of the image (128 pixels) to predict the second half (128 pixels) of the image.

\begin{figure}[h]
\centering
\begin{minipage}[b]{0.48\linewidth}
{\epsfig{file = 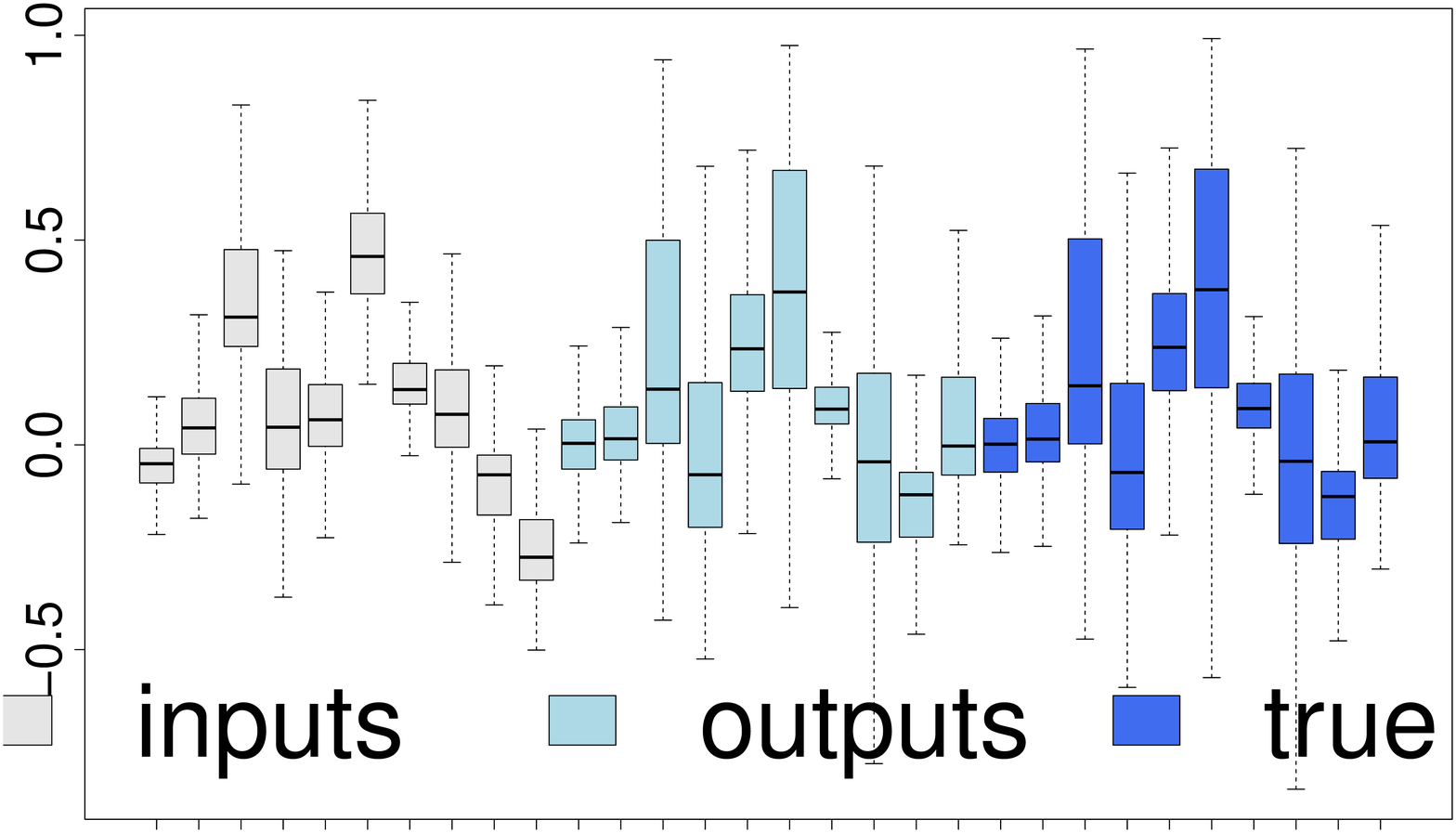, width = 7.005cm,height = 4cm}}
  \captionof{figure}{Effect of the auto-encoder: transforming input (gray) to output (light blue).}
  \label{fig:auto}
  \end{minipage}
\quad
\begin{minipage}[b]{0.48\linewidth}
{\epsfig{file = 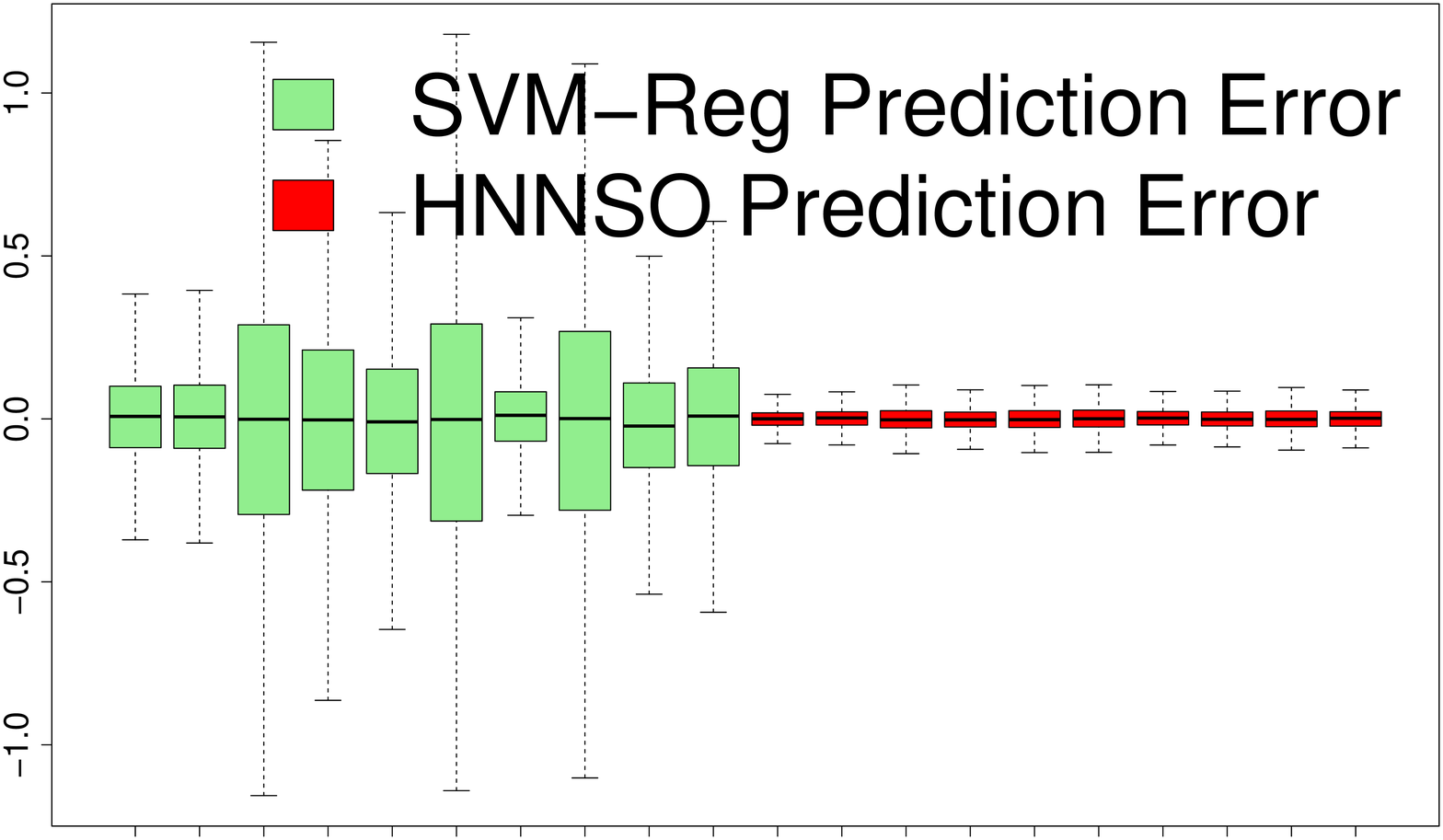, width = 7.005cm,height = 4cm}}
  \captionof{figure}{Errors made by the SVM-Reg (green) and HNNSO (red) for each target.}
  \label{fig:errdistribution} 
  \end{minipage}
\end{figure}

\begin{figure}[h]
\centering
{\epsfig{file = 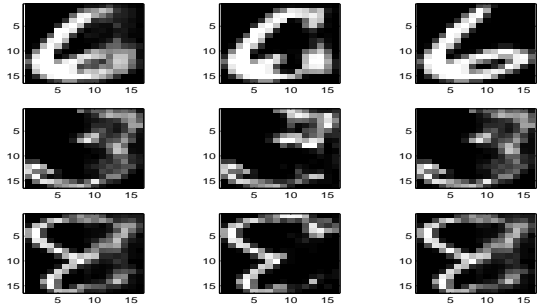, width = 7.105cm,height = 4.0482480000112395670016cm}}
  \captionof{figure}{Predicting the right half of a digit using the left half in the USPS data}
  \label{fig:usps} 
  \end{figure}

\textbf{General Performance} \newline
Table \ref{tab:all} presents the performance of different 
regression models on the SSTB, MNIST, and USPS datasets. The results show that the HNNSO achieves significantly lower RMSE scores in comparison to other models. On all three datasets, the relative error reduction achieved by HNNSO over other methods was at least 6.4\% (ranging between 6.4\% and 54.9\%).  

\textbf{Detailed Anaylsis} \newline
We use the SSTB dataset to gain some insights into the HNNSO's modeling behavior.  Performance-wise, we have shown above that the HNNSO model achieved a RMSE score of 0.0373 on the SSTB data. Without pretraining,  the error increases relatively by 9.4\%. 
Figure~\ref{fig:dist} further depicts the distribution of the first output variable of the data. The figure indicates that the distribution of the input with pretraining (middle), compared to that without pretraining (left), is closer to the distribution of the true labels (right). 
Such structured patterns are important for the encoder as discussed earlier.

In Figure~\ref{fig:auto}, we also show the input (gray boxes) and output (light-blue) of the auto-decoder in HNNSO as well as  the true labels (dark-blue) on the SSTB data. Each box in each color group represents one of the ten output variables in the same order. Figure~\ref{fig:auto} shows that the patterns of the light-blue boxes are similar to that of the dark-blue boxes. This suggests that the encoder is able to guide the output predictions to follow similar structured patterns as that of the true labels. 

In Figure~\ref{fig:errdistribution}, we further depict the errors made by the HNNSO and SVM-Reg (the second best approach). Each box in each color group represents the error, calculated as predicted value minus its true value, achieved on each of the ten output variables in the same order.  Figure~\ref{fig:errdistribution} suggests that the errors on each output target made by HNNSO has narrow and consistent variances across the ten output targets. On the contrary, the variances of errors among the ten output targets obtained by the SVM-Reg are obviously larger, suggesting that SVM-Reg makes good prediction on some output targets without considering the interactions with other targets.  

\textbf{Visualization} \newline
Figure~\ref{fig:usps} plots three digits from the USPS data, including the true images (right) and their predictions made by HNNSO (left) and MLP (middle).  The figure shows that HNNSO was able to recover the images well. 
 In contrast, MLP yielded some missing pixels on the right halves of the images.

\section{Conclusion}
We propose a deep high-order neural network to construct nonlinear functional mappings from structured input to structured output for regression. We aim to jointly achieve the goal with complementary components that focus on capturing different types of interdependency.  
Experimental results on three benchmarking datasets show the advantage of our model over several competing approaches. In the future, we plan to explore our strategy with a hinge loss for structured label classification with applications in image labeling and scene understanding.



\bibliographystyle{abbrv}

\bibliography{references}


\end{document}